\documentclass[10pt,twocolumn,letterpaper]{article}

\usepackage[pagenumbers]{iccv} 
\definecolor{deepblue}{HTML}{27a2c3}

\definecolor{iccvblue}{rgb}{0.21,0.49,0.74}
\usepackage[pagebackref,breaklinks,colorlinks,allcolors=iccvblue]{hyperref}

\def\paperID{*****} 
\def\confName{ICCV}
\def\confYear{2025}

\title{ChatVLA: Unified Multimodal Understanding and Robot Control \\with Vision-Language-Action Model}

\author{
Zhongyi Zhou$^{1,2*}$ \quad Yichen Zhu$^{1*\dagger}$\quad Minjie Zhu$^{2}$\quad Junjie Wen$^{2}$\quad Ning Liu$^{4}$ \quad Zhiyuan Xu$^{4}$\\ 
Weibin Meng$^{5}$\quad Ran Cheng$^{1}$\quad Yaxin Peng$^{3}$\quad Chaomin Shen$^{2}$\quad Feifei Feng$^{1}$ \vspace{0.03in} \\
$^1$Midea Group \quad $^2$East China Normal University \quad $^3$Shanghai University
\\ \quad $^4$ Beijing Innovation Center of Humanoid Robotics \quad $^5$Tsinghua University \\
\thanks{$*$ Co-first author. $\dagger$ Corresponding author.}
\vspace{-0.1in} \\
\href{https://chatvla.github.io}{\color{deepblue}\textbf{chatvla.github.io}\xspace} \vspace{-0.3in}
}

\begin{document}

\makeatletter
\let\@oldmaketitle\@maketitle%
\renewcommand{\@maketitle}{\@oldmaketitle
    \begin{center}
        \captionsetup{type=figure}
        \centering
    \includegraphics[width=1.02\textwidth]{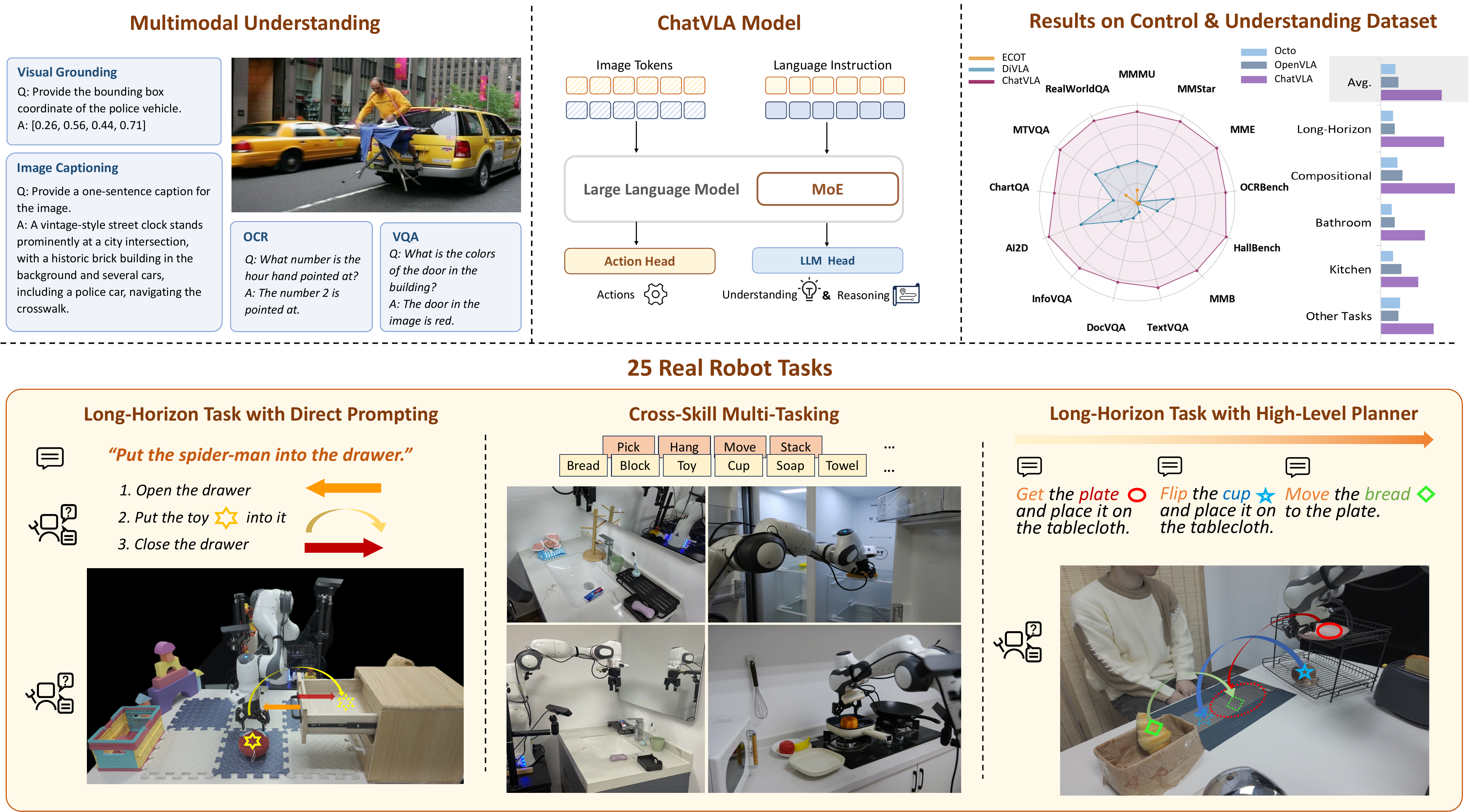}
    \caption{\textbf{ChatVLA is the first work to unify multimodal understanding and embodied control.} We conduct extensive evaluations on VQA and multimodal understanding benchmarks to demonstrate that robot foundation models can also engage in chat. Furthermore, we evaluate our approach on diverse real-world robot tasks.}
    \label{fig:firstfig}
    \end{center}
}
\def\thanks#1{\protected@xdef\@thanks{\@thanks
        \protect\footnotetext{#1}}}
\makeatother
\maketitle
\begin{abstract}
Humans possess a unified cognitive ability to perceive, comprehend, and interact with the physical world. Why can't large language models replicate this holistic understanding? Through a systematic analysis of existing training paradigms in vision-language-action models (VLA), we identify two key challenges: \textit{spurious forgetting}, where robot training overwrites crucial visual-text alignments, and \textit{task interference}, where competing control and understanding tasks degrade performance when trained jointly. To overcome these limitations, we propose ChatVLA, a novel framework featuring Phased Alignment Training, which incrementally integrates multimodal data after initial control mastery, and a Mixture-of-Experts architecture to minimize task interference. ChatVLA demonstrates competitive performance on visual question-answering datasets and significantly surpasses state-of-the-art vision-language-action (VLA) methods on multimodal understanding benchmarks.  Notably, it achieves a six times higher performance on MMMU and scores 47.2\% on MMStar with a more parameter-efficient design than ECoT. Furthermore, ChatVLA demonstrates superior performance on 25 real-world robot manipulation tasks compared to existing VLA methods like OpenVLA. Our findings highlight the potential of our unified framework for achieving both robust multimodal understanding and effective robot control.
\end{abstract}
\section{Introduction}
Recent advancements in Vision-Language-Action (VLA)~\cite{[pi0, kim24openvla, wen2024tinyvla, wen2025dexvla} models have largely prioritized robotic action mastery. While models trained on robotic control tasks excel at low-level manipulation and physical interaction, they often struggle to interpret and reason about multimodal data like images and text. This is paradoxical, as modern VLA architectures build upon pre-trained vision-language models (VLMs). Conversely, VLMs trained on visual-text pairs demonstrate impressive multimodal scene understanding but lack the ability to physically interact with the environment. This duality highlights a critical challenge: unifying embodied control and multimodal understanding by aligning these disparate data sources (robotic actions and visual-text semantics) without sacrificing performance in either domain.

This work investigates how to unify a single end-to-end neural network capable of multimodal scene understanding, conversational ability, and physical interaction. We first explore existing training paradigms to assess their feasibility for unification. Specifically, we examine three data settings for VLA training: 1) training solely on expert demonstration data containing robot action trajectories (the most common approach, e.g., OpenVLA~\cite{kim24openvla}, TinyVLA~\cite{wen2024tinyvla}, $\pi_0$~\cite{[pi0}); 2) augmenting robot data with reasoning phrases to guide action (similar to ECoT~\cite{ecot} and DiffusionVLA~\cite{diffusionvla}); and 3) co-training with both visual-text pairs and robot data (as in RT-2~\cite{rt-2}). We analyze how each configuration impacts the model's ability to balance control and understanding. Our experiments reveal that training solely with robot data erodes conversational ability entirely; adding reasoning data partially preserves multimodal understanding; and introducing visual-text pairs significantly weakens control capabilities.  This suggests two key challenges: (1) VLA models suffer from \textbf{spurious forgetting}~\cite{zheng2025spurious, zhai2023investigating, luo2023empirical}, where performance degradation may not reflect complete knowledge loss from pre-trained VLMs, but rather a shift in how the model aligns its internal representations with different tasks. The alignment between robot actions and visual-text data appears fragile and susceptible to being overwritten during fine-tuning. (2) \textbf{Task interference}~\cite{wang2021afec,ahn2025prevalence} arises, where the conflicting parameter spaces of control and understanding tasks, sharing overlapping representations, cause mutual performance degradation when trained simultaneously. 

To address these challenges, we present ChatVLA, a simple yet effective framework—in terms of both neural architecture and training strategy—for enabling a single neural network to master both understanding and manipulation. We propose Phased Alignment Training, a two-stage strategy inspired by curriculum learning. The model first masters embodied control before incrementally integrating multimodal data to ``reactivate" frozen alignment links. Furthermore, we introduce a Mixture-of-Experts (MoE) on the MLP layers. This allows the two tasks to share attention layers (for cross-task knowledge transfer) while isolating task-specific MLPs (to minimize interference). This design is motivated by Dual Coding Theory~\cite{paivio1991dual}, which posits that human minds process information through two separate but interconnected systems: one for physical skills and the other for verbal and visual practice. The shared attention layers in ChatVLA facilitate the exchange of mutually beneficial knowledge between understanding and control tasks, while the separate MLP layers process learned knowledge independently.

We evaluate ChatVLA across three dimensions: conversational ability (visual question answering), general multimodal understanding, and general robot control. Specifically, we assess its conversational ability on established datasets like TextVQA~\cite{textvqa} and DocVQA~\cite{mathew2021docvqa}, where it achieves competitive performance compared to existing VLMs. Furthermore, ChatVLA demonstrates strong multimodal understanding capabilities on general visual and textual benchmarks, including MMMU~\cite{yue2023mmmu}, MME~\cite{mme}, and MMStar~\cite{chen2024we}. Notably, compared to state-of-the-art VLA methods like ECoT, our method achieves a 6x performance improvement on MMMU and boosts performance on MMStar from 0 to 47.2, using 3.5x fewer parameters in the VLM backbone. Finally, we evaluate ChatVLA on 25 real-world robot tasks encompassing diverse skills like picking, placing, pushing, and hanging, across multiple environments such as bathrooms, kitchens, and tabletops. In this multi-task setting, our method outperforms state-of-the-art VLA methods like OpenVLA. These results validate the effectiveness of our approach, showcasing the potential of a single unified method for both multimodal understanding and robot control.

In summary, our contributions are the following:
\begin{itemize}
    \item We provide an in-depth analysis of existing VLA approaches under rigorous settings, demonstrating their limitations in achieving satisfactory performance across both multimodal understanding and robot control.
    \item We introduce ChatVLA, a simple yet effective framework that unifies conversational ability, multimodal understanding, and robot control within a single neural network.
    \item We conduct extensive experiments to evaluate ChatVLA's performance on various question-answering and general understanding benchmarks. 
    \item We perform extensive real-world robot experiments, encompassing 25 diverse tasks in realistic home environments (tabletop, kitchen, and bathroom), demonstrating ChatVLA's superior performance in real-world robot control scenarios.
\end{itemize}

\section{Related Work}

\textbf{Multimodal understanding}
Multimodal Large Language Models (MLLMs)~\cite{lu2024deepseek-vl, openflamingo, idefics, llava, llava1.5, wang2024qwen2, chen2024internvl, zhu2024comprehensive, ma2024janusflow, zhou2024transfusion, minigpt4, luo2024mono, chen2024internvl, blip-2, instructblip, chen2024expanding,karamcheti2024prismatic} have significantly advanced the field of multimodal understanding by integrating visual and linguistic information to achieve holistic scene comprehension. MLLMs have demonstrated excellent performance on tasks requiring cross-modal alignment, such as visual question answering (VQA), image captioning, and spatial reasoning. This success stems from their ability to map visual features to semantic representations through sophisticated adapter designs. However, current MLLMs lack a connection to the physical world, preventing them from interacting with environments and humans. This work aims to bridge this gap, enabling vision-language models to also act.

\textbf{Vision-langauge-action models in robot learning.} Vision-language-action models (VLAs) form a growing body of research that leverages pre-trained vision-language models (VLMs) as a backbone to enable both language comprehension and observational understanding. These methods typically fine-tune large pre-trained VLMs to predict robot actions~\cite{brohan2023rt-2, roboflamingo, leo3d, wen2024tinyvla, pertsch2025fast, [pi0, kim24openvla, diffusion-policy, zhu2024scalingdp, wang2024sparse-dp, prasad2024consistencypolicy, black2023training, black2023zero, dasari2024ingredients, lin2024datascalinglawsimitation, multimodal_diffusion_transformer, aloha_unleashed, uehara2024fine, uehara2024feedback,ding2024quar}. These methods have shown strong performance in both simulated and real-world tasks. However, existing VLA models have not demonstrated the ability to perform true multimodal understanding. Based on our experiments, we find that these models lack this capability. In contrast, our work proposes a unified approach that enables a single network to effectively handle both multimodal understanding and robot control.
\begin{figure*}[t]
    \centering
    \includegraphics[width=.9\linewidth]{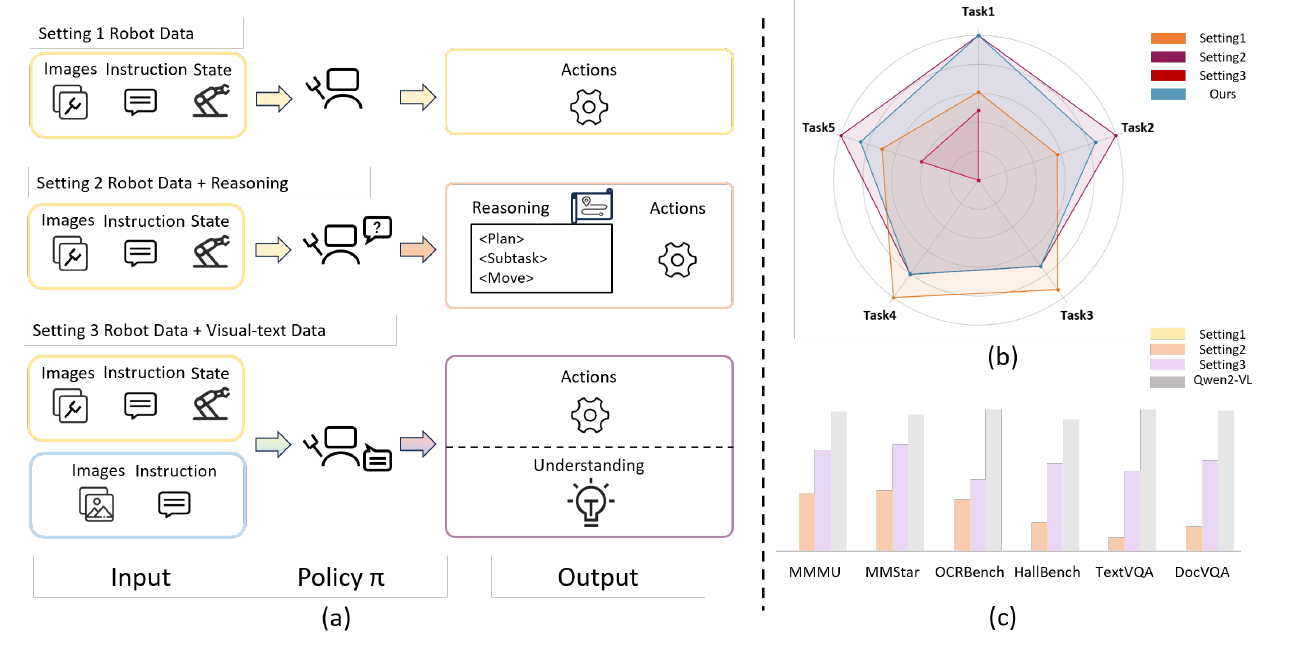}
    \caption{\textbf{Analysis of how training data influences VLA performance on control and understanding tasks.} (a) We use three different sets of training data, corresponding to the three main training approaches for VLA models. (b) The experimental results are presented for five real-world robot tasks across three settings. (c) The results on VQA and multimodal understanding benchmarks.}
    \label{fig:setting}
\end{figure*}

\section{Methodology}

This section provides a thorough discussion of our framework. Section~\ref{sec:definition} presents formal definitions. Section~\ref{sec:motivation} details our motivation and empirical results demonstrating how existing vision-language-action models (VLAs) suffer from catastrophic forgetting and task interference, thus hindering the unification of multimodal understanding and robot control. Section~\ref{sec:architecture_data} proposes a simple solution to address this problem.

\subsection{Formal Definition}\label{sec:definition}
Consider two distinct scenarios: robot control and multimodal understanding. In the context of robot control, we typically construct a dataset of demonstrations $D_{robot} = \{\tau_{i}\}_{i=1}^{N}$, where each demonstration $\tau_{i}$ comprises a sequence of state-action pairs.  The state $s$ consists of an observation (image) $v$ and an instruction (text) $t$, such that $s = (v, t)$.  We can represent the sequence of state-action pairs as $\tau_{i} = \{((v_{1},t_{1}),a_{1}), ((v_{2},t_{2}),a_{2}), \dots, ((v_{T},t_{T}),a_{T})\}$, where each tuple $((v_j, t_j), a_j)$ represents the state at timestep $j$ and the corresponding action taken, and $T$ is the length of the demonstration. These demonstrations are typically provided by a human expert.

For multimodal understanding and visual conversation tasks, we have a dataset $D_{v-t} = \{\phi_{i}\}_{i=1}^{M}$, where each data sample $\phi_{i}$ consists of a visual image $v_i$ and a corresponding question (or caption) in textual form $t_i$, i.e., $\phi_{i} = \{(v_i, t_i)\}$. Here, $M$ represents the total number of such image-text pairs. The notation $v-t$ denote visual-text data.

The overarching goal of our work is to develop a general model $\pi$ capable of addressing both embodied control and multimodal understanding.  For embodied control, this involves learning a policy that models the joint distribution of robot actions given the current visual observation and textual instruction: $\pi(a_{t}|v_{t},t_{t})$.  Simultaneously, for multimodal understanding and visual question answering, the model should capture the distribution of the text (answer or caption) given the visual input: $\pi(t|v)$.  Our objective is to create a unified model that can effectively learn both distributions, enabling it to perform well in both robot control tasks and multimodal understanding scenarios.

Current VLA research focuses on developing more robust and generalizable models for learning visuomotor policies~\cite{kim24openvla,[pi0,wen2024tinyvla}. Some approaches explore chain-of-thought-like reasoning to improve policy generation~\cite{ecot,diffusionvla,coa-vla}, while others investigate co-training VLA models with visual-text and robot data~\cite{pertsch2025fast}. In particular, some studies report benefits from co-training with visual-text data in laboratory settings~\cite{rt-2}, while others find it less effective in real-world scenarios~\cite{ecot}. Although a few works suggest that VLA can maintain conversational ability~\cite{diffusionvla,rt-2}, none have thoroughly investigated how this ability, along with general multimodal understanding, is preserved after applying the VLA training paradigm. In the following section, we analyze different training data setups for VLA, focusing specifically on the resulting model's performance in both multimodal understanding and real-world robot control. Our goal is to provide practical guidance for building unified models capable of both.

\subsection{Analysis}\label{sec:motivation}
To understand the capabilities of existing VLA models in terms of multimodal understanding and embodied control, we investigate three distinct training paradigms, each utilizing a different dataset: 1) training solely with robot data, the most prevalent approach in VLA~\cite{[pi0,openflamingo,kim24openvla,wen2024tinyvla}, primarily focused on optimizing robot control performance; 2) augmenting robot data with chain-of-thought-like reasoning, aiming to provide auxiliary information that improves both model generalization and robot task performance~\cite{diffusionvla,ecot}; and 3) co-training with both visual-text data and robot data. This latter paradigm was pioneered by RT-2~\cite{rt-2}; however, due to proprietary data and model details, exact replication is challenging. Following RT-2, we used a 3:1 ratio of robot data to visual-text data in this experiment.

In this section, we analyze these three training data setups for VLA models. Specifically, we utilize DiffusionVLA~\cite{diffusionvla}, a representative VLA model that supports both language output via autoregression and action generation via a diffusion model. We evaluate performance on six representative benchmarks: four focused on visual question answering and two providing a broader evaluation of multimodal large language models, encompassing tasks like math and OCR.  Furthermore, we assess performance on five real-world robot tasks covering diverse skills, including hanging, pulling, picking, and placing.  Following the methodology of DiffusionVLA, we generate robot reasoning data. For visual-text data, we randomly sample 54k image-text pairs from LLaVA. Further details regarding experimental setup and data processing are provided in the Appendix.

\textbf{Results on multimodal understanding and question-answering benchmark.} The experimental results are presented in Figure 2. The bottom-right portion of the figure displays performance on six benchmarks, encompassing both visual question answering (VQA) and general understanding tasks. The top-right portion of Figure 2 shows the average success rate across a total of 112 trials conducted on five real-world robot tasks. 

The bottom-right table includes results for the base model, Qwen2-VL~\cite{wang2024qwen2}. Some results are intuitive. For example, training the model solely on robot data yields a performance of 0 across all benchmarks. This model completely loses its conversational ability, exhibiting only murmuring when asked a question. As expected, the smallest performance drop compared to the base model occurs when training uses both visual-text pairs and robot data. Interestingly, training with robot data including reasoning also boosts performance from 0 to a non-negligible level, despite the highly structured, template-driven nature of the reasoning phrases within that data. Even though the reasoning phrases are similar and structured, explicitly allowing the model to ``speak out" significantly improves performance on question answering and even general understanding.

\textbf{\textit{Conclusion 1.}} Our observations suggest that the pre-trained VLM component suffers from what appears to be catastrophic forgetting. Training solely with robot data causes the model to lose previously acquired conversational and understanding abilities. However, our experiments indicate that this isn't necessarily a complete loss of knowledge, but rather a misalignment caused by the robot data. Training with a fixed reasoning template seems to ``reactivate" the visual-text alignment, enabling the model to engage in conversation and demonstrate understanding. In Section~\ref{sec:appendix}, we will delve into the specific knowledge that is reactivated and discuss how future work can further bridge the gap between the base VLM and the VLA model. We term this phenomenon ``spurious forgetting."

\begin{figure}[t]
    \centering
    \includegraphics[width=1\linewidth]{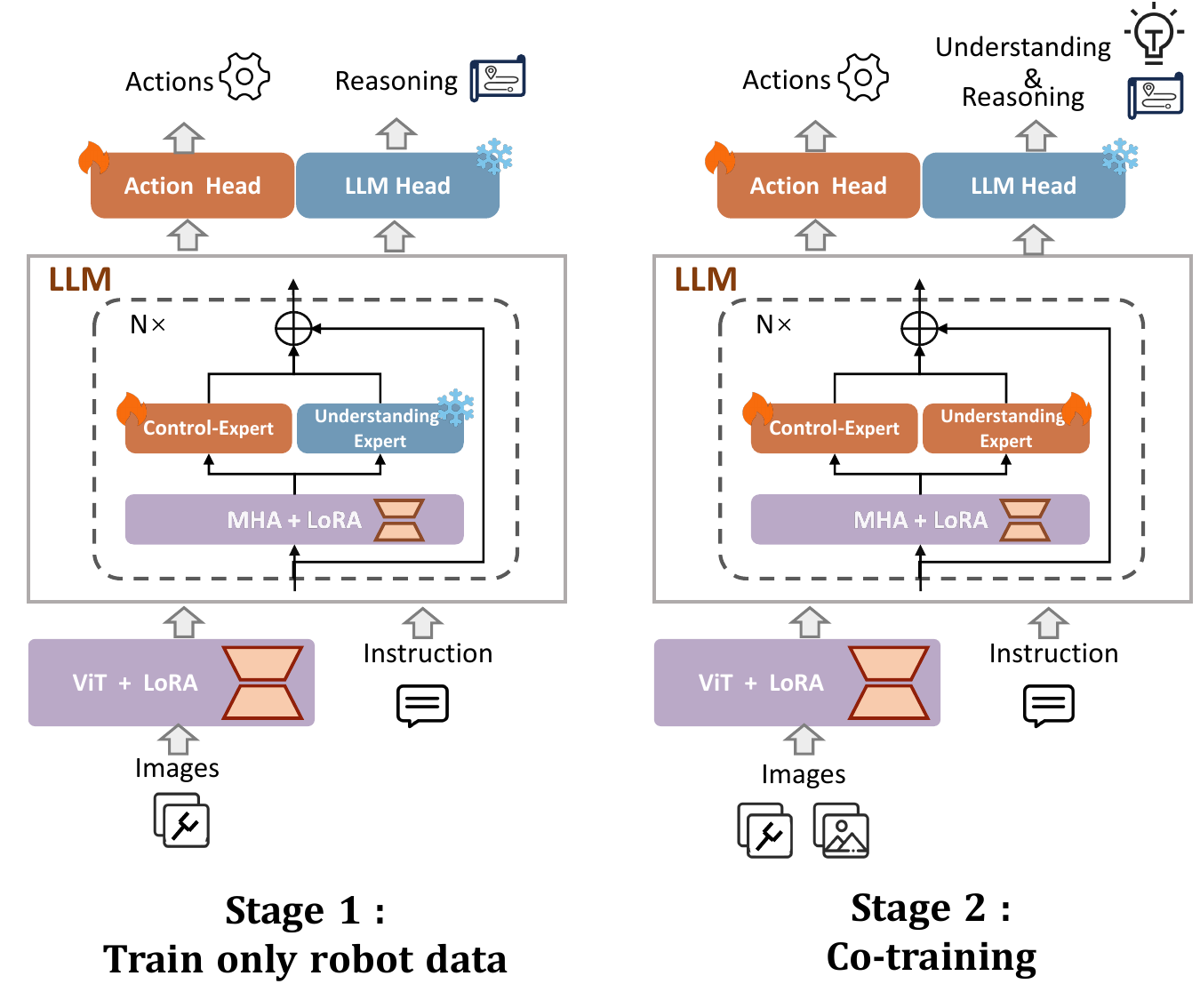}
    \caption{\textbf{Training strategy.} Our framework is initially trained on robot data with action trajectories, then co-trained with visual-text and robot data to maintain performance in both domains.}
    \label{fig:stage}
\end{figure}
\textbf{Results on real robot multi-task settings.} We further evaluated different approaches to our real robot setup. All methods were trained on 25 real robot tasks, and we selected five diverse tasks, covering skills like pushing, picking, and hanging, for comparison.  Details, including the number of trials for each experiment, can be found in the Appendix.  Surprisingly, training with only robot data yielded worse performance than incorporating reasoning. This confirms previous findings that leveraging either visual or textual chain-of-thought enhances the generalization of robot models.  Intriguingly, co-training robot data with visual-text data resulted in a significant performance drop in real-world task success rates.

\textbf{\textit{Conclusion 2.}} The initial observation that incorporating reasoning into robot data improves performance aligns with Dual Coding Theory~\cite{paivio1991dual}.  This theory posits that physical motor skills and visual-linguistic understanding are not mutually exclusive but rather interconnected, offering overlapping benefits. However, the performance of robot control dramatically decreased when visual-text pairs were added to the training data. This suggests that the distinct representations required for action generation and understanding may compete within the shared parameter space. This phenomenon, we named as \textbf{partial task interference}, requires careful resolution. A unified system should connect the two data types while simultaneously enabling separable representation learning for each task.

\begin{figure}[t]
    \centering
    \includegraphics[width=.9\linewidth]{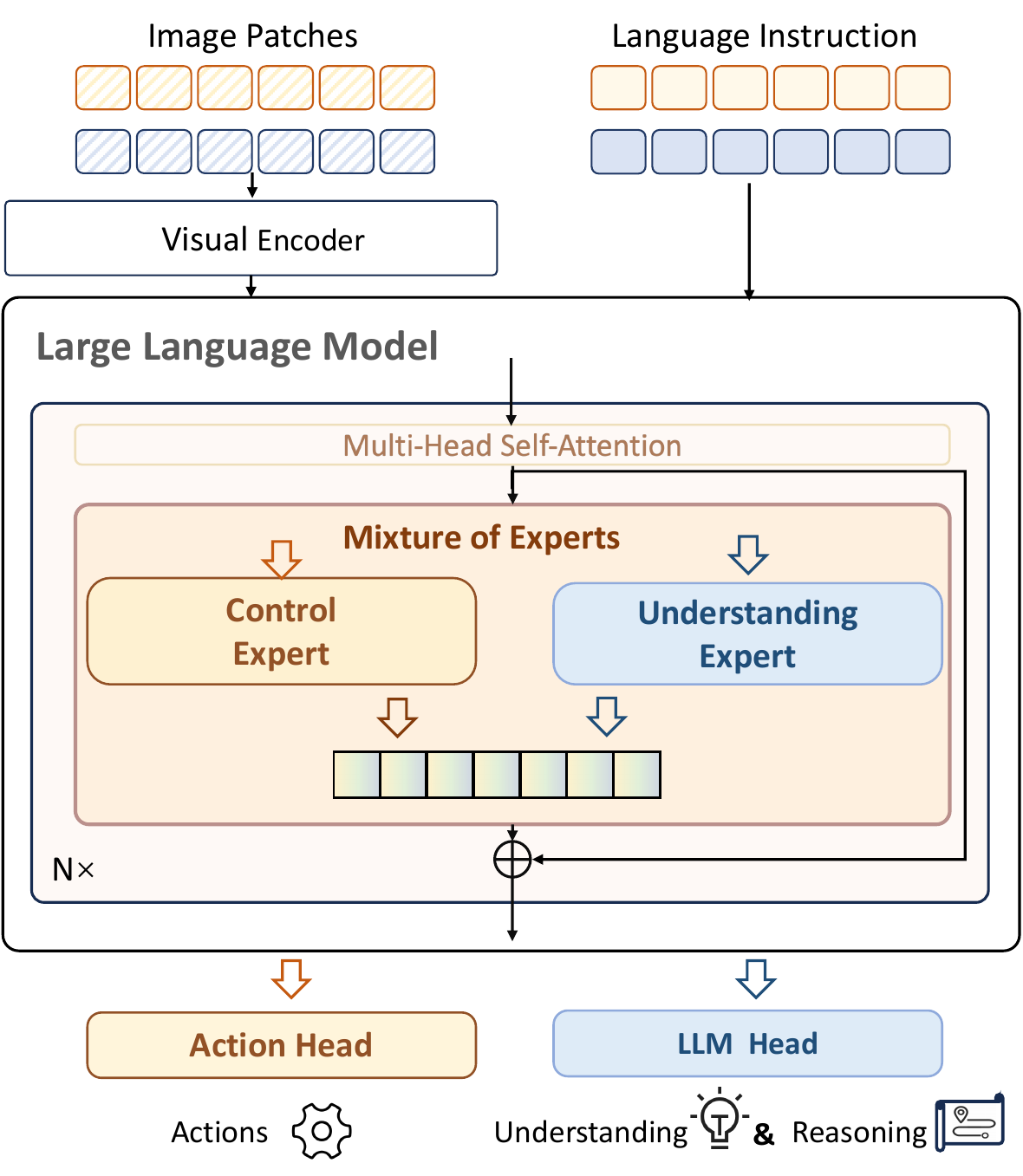}
    \caption{\textbf{Illustration of the Mixture-of-Experts component of ChatVLA.} Two distinct expert types process robot data and visual-text data separately, while shared self-attention layers facilitate knowledge transfer between the two domains.}
    \label{fig:framework}
    \vspace{-1mm}
\end{figure}

\begin{table*}[tb]
  \centering
  \caption{\textbf{Understanding task:} Evaluation of MLLMs and VLAs on 6 Multimodal Understanding benchmarks and 7 VQA benchmarks. The bold denotes top-ranked methods, underlined entries signify secondary performers.} 
  \label{tbl:qa_table}
  \resizebox{1.0\linewidth}{!}{
      \begin{tabular}{c|c|cccccc|ccccccc}
        \toprule
        \multirow{2}{*}{Method} &\multirow{2}{*}{\#Params} & \multicolumn{6}{c|}{Multimodal Understanding Benchmarks} & \multicolumn{7}{c}{VQA Benchmarks} \\
               & & MMMU & MMStar & MME & OCRBench & HallBench & MMB 
                & TextVQA & DocVQA & InfoVQA & AI2D & ChartQA & MTVQA & RealWorldQA \\
        \midrule
        \multicolumn{15}{c}{\textbf{Multimodal Large Language Models}} \\
        \midrule
        Janus~\cite{wu2024janus} & 1.3B & 30.5 & 37.6 & 1338.0 & 482 & 30.3 & 69.4 & — & — & — & 52.8 & — & — & — \\
        DeepSeek-VL~\cite{lu2024deepseek-vl} & 1.3B & 32.2 & 39.9 & — & 409 & 27.6 & 64.6 & — & — & — & 51.5 & — & — & — \\
        Qwen2-VL~\cite{wang2024qwen2}       & 2B & \textbf{41.1} &  \textbf{48.0} &  \textbf{1872.0} & \textbf{809} &  \textbf{41.7} & \underline{74.9} &  \textbf{79.7} &  \textbf{88.57} &  \textbf{61.37} & \underline{74.7} & \underline{73.5} &  \textbf{18.1} &  \textbf{62.9} \\
        SmolVLM~\cite{huggingface2023smolvlm} & 2.3B & 38.8 & 41.7 & — & 656 & 39.5 & — & \underline{72.7} & 81.6 & — & 64.2 & — & — & — \\
        LLaVA-Phi~\cite{llavaphi} & 2.7B & — & — & 1335.1 & — & — & 59.8 & 48.6 & — & — & — & — & — & — \\
        MobileVLM-V2~\cite{mobilevlmv2} & 3B & — & — & 1440.5 & — & — & 63.2 & 57.5 & — & — & — & — & — & — \\
        MoE-LLaVA~\cite{moe-llava} & 3.6B & — & — & 1431.3 & — & — & 68 & 57 & — & — & — & — & — & — \\
        Phi-3-Vision~\cite{abdin2024phi} & 4.2B& \underline{40.4} & — & — & — & — & 80.5 & 70.9 & — & — &  \textbf{76.7} &  \textbf{81.4} & — & — \\
        LLaVA-1.5~\cite{llava1.5} & 7B & 34.2 & — & \underline{1510.7} & — & — & 64.3 & 58.2 & — & — & 63.1 & 55.0 & — & — \\
        DeepSeek-VL~\cite{lu2024deepseek-vl} & 7B & 36.6 & — & — & 456 & — & 73.2 & — & — & — & — & — & — & — \\
        LLaVA-Next~\cite{li2024llavanext-strong} & 8B & 36.4 & — & — & — & — &  \textbf{79.7} & 55.7 & — & — & 66.9 & 65.8 & — & — \\
        \midrule
        \multicolumn{15}{c}{\textbf{Vision-Language-Action Models}} \\
        \midrule
        OpenVLA~\cite{kim24openvla} & 7B & 0 & 0& 0& 0& 0& 0& 0& 0& 0 & 0& 0& 0& 0\\
        ECoT~\cite{ecot}     & 7B & 5.4 & 0 & 0 & 12 & 0.9 & — & 0 & 0 & 0 & 0 & 0 & 1.7 & 0 \\
        DiVLA~\cite{diffusionvla} & 2B & 17.2 & 21.1 & 186.5 & 294 & 9.0 & — & 7.5 & 15.2 & 14.7 & 43.1 & 17.2 & 6.2 & 25.2\\
       \textbf{ChatVLA(Ours)} & 2B & \textbf{37.4} & \textbf{\underline{47.2}} & \textbf{1435.2} & \textbf{\underline{729}} & \textbf{\underline{39.9}} & \textbf{69.0} & \textbf{71.2} & \textbf{\underline{83.3}} & \textbf{\underline{53.3}} & \textbf{67.6} & \textbf{59.9} & \textbf{\underline{11.5}} & \textbf{\underline{57.0}} \\
        \bottomrule
      \end{tabular}
  }
\end{table*}

\subsection{Method: ChatVLA}
\label{sec:architecture_data}
As discussed above, training on robot policy data can interfere with learning of visual-text relationships.  Furthermore, training exclusively on robot data can diminish visual-text alignment, leading to a degradation of the model's conversational abilities. Therefore, addressing these two challenges is crucial for successfully unifying both perspectives within a single VLA model. We will first describe the training strategy used to address spurious forgetting, and then outline the general architecture of our method to tackle the second challenge.

\textbf{Phased alignment training.} Previously, we identified that spurious forgetting is a key factor in causing VLA to lose its ability to chat and understand scenes. Since the pre-trained VLM is well-trained and excels at visual-related tasks, it is intuitive that the ability to chat and understand scenes can be reactivated with a small amount of visual-text pair data. In contrast, robot control tasks are much more complex to train, so the priority should be to develop an excellent model that excels at embodied control tasks. Our training strategy is straightforward yet effective. We first train the VLA model on robot data. During this training, we also include reasoning data to ensure continuous alignment between the visual and text components. Once the robot data is trained, we co-train both visual-text and robot data to help the model retain proficiency in both tasks.

\begin{table*}[tb]
  \centering
  \caption{\textbf{Long-horizon real robot tasks with direct prompting.} \textit{The task is completed in a sequence.} The Avg. Len. denotes the average success length of the model. Task 1: Sort toys. Task 2: Stack building blocks. Task 3: Place the toy in the drawer. Task 4: Clean building blocks to the box.}

  \label{tbl:taskscore_longhorizon}
  \resizebox{.9\linewidth}{!}{
      \begin{tabular}{c|cccc|c|cc|c|ccc|c|cc|c}
        \toprule
        \multirow{2}{*}{Method} & \multicolumn{5}{c|}{Task 1} & \multicolumn{3}{c|}{Task 2} & \multicolumn{4}{c|}{Task 3} & \multicolumn{3}{c}{Task 4} \\
        \cmidrule(lr){2-6} \cmidrule(lr){7-9} \cmidrule(lr){10-13} \cmidrule(lr){14-16}
        & 1 & 2 & 3 & 4 & Avg. Len. & 1 & 2 & Avg. Len. & 1 & 2 & 3 & Avg. Len. & 1 & 2 & Avg. Len.\\ 
        \midrule
         
    Octo~\cite{octo} & 0.23 & 0.08 & 0.00 & 0.00 & 0.08 & 0.29 & 0.14 & 0.21 & 0.11 & 0.11 & 0.11 & 0.11 & 0.50 & 0.17 & 0.33\\
    OpenVLA~\cite{kim24openvla} & 0.15 & 0.08 & 0.00 & 0.00 & 0.06 & 0.43 & 0.14 & 0.29 & 0.22 & 0.11 & 0.11 & 0.15 & 0.50 & 0.33 & 0.42\\
   \textbf{ChatVLA(Ours)} & \textbf{0.92} & \textbf{0.69} & \textbf{0.31} & \textbf{0.23} & \textbf{0.54} & \textbf{0.86} & \textbf{0.43} & \textbf{0.64} & \textbf{1.00} & \textbf{1.00} & \textbf{1.00} & \textbf{1.00} & \textbf{0.83} & \textbf{0.67} & \textbf{0.75}\\
        \bottomrule
      \end{tabular}
    }
\end{table*}

\begin{table*}[t]
  \centering
  \caption{\textbf{Long-horizon real robot tasks with high-level policy model.} \textit{The task is completed in a sequence.} The Avg. Len. denotes the average success length of the model.  Task 5-8: Move the block to the basket then put the toy into the drawer. Task 9-10: Move two blocks to the basket sequentially. Task 11-13: Prepare the breakfast for me. }

  \label{tbl:taskscore_compositional}
  \resizebox{0.85\linewidth}{!}{
      \begin{tabular}{c|cccc|c|cc|c|ccc|c}
        \toprule
        \multirow{2}{*}{Method} & \multicolumn{5}{c|}{Task 5-8} & \multicolumn{3}{c|}{Task 9-10} & \multicolumn{4}{c}{Task 11-13} \\
        \cmidrule(lr){2-6} \cmidrule(lr){7-9} \cmidrule(lr){10-13} 
        & 1 & 2 & 3 & 4 & Avg. Len. & 1 & 2 & Avg. Len.& 1 & 2 & 3 & Avg. Len. \\
        \midrule
        Octo~\cite{octo} & 0.42 & 0.25 & 0.17 & 0.08 & 0.23 & 0.33 & 0.22 & 0.28 & 0.15 & 0.08 & 0.00 & 0.08\\
        OpenVLA~\cite{kim24openvla} & 0.42 & 0.33 & 0.33 & 0.17 & 0.31 & 0.44 & 0.22 & 0.33 & 0.23 & 0.08 & 0.00 & 0.10\\
        \textbf{ChatVLA(Ours)} & \textbf{1.00} & \textbf{0.92} & \textbf{0.92} & \textbf{0.92} & \textbf{0.94} & \textbf{0.89} & \textbf{0.78} & \textbf{0.83} & \textbf{0.69} & \textbf{0.54} & \textbf{0.54} & \textbf{0.59}\\

        \bottomrule
      \end{tabular}
    }
\end{table*}


\begin{table*}[tb]
  \centering
  \caption{\textbf{Real robot multi-tasking.} We evaluated our model in a multi-task setting across diverse scenes, including bathrooms, kitchens, and tabletops. The Avg. denotes the average success rate. These tasks also encompassed a range of skills. }
  \label{tbl:taskscore_other}
  \resizebox{1.0\linewidth}{!}{
      \begin{tabular}{c|cccc|cc|cccccc|c}
        \toprule
        \multirow{2}{*}{Method} & \multicolumn{4}{c|}{Bathroom} & \multicolumn{2}{c|}{Kitchen} & \multicolumn{6}{c|}{Tabletop} & \multirow{2}{*}{Avg} \\
        \cmidrule(lr){2-5} \cmidrule(lr){6-7} \cmidrule(lr){8-13} 
        &  Task 14 & Task 15 & Task 16 & Task 17 &Task 18 & Task 19 &Task 20 & Task 21 & Task 22 & Task 23 & Task 24 & Task 25 & \\
        \midrule
         Octo~\cite{octo} & 3/11 & 0/6 & 1/9 & 0/7 &  0/11 & 3/11 & 1/7 & 2/9 & 1/7 & 2/13 & 2/9 & 3/7  & 18/107 \\
        OpenVLA~\cite{kim24openvla} &  2/11 & 0/6 & 2/9 & 1/7 &  1/11 & 4/11  & 2/7 & 1/9 & 1/7 & 4/13 & 0/9 & 2/7 &  20/107 \\
         \textbf{ChatVLA(Ours)} & \textbf{6/11} & \textbf{2/6} & \textbf{5/9} & \textbf{3/7} &  \textbf{3/11} & \textbf{6/11} & \textbf{4/7} & \textbf{5/9} & \textbf{4/7} & \textbf{6/13} & \textbf{4/9} & \textbf{7/7} & \textbf{55/107} \\
        \bottomrule
      \end{tabular}
    }
\end{table*} 

\textbf{Mixture of experts.} The previous section demonstrated the use of phased alignment training to address the spurious forgetting problem, enabling the model to retain knowledge from the previously trained VLM. However, this approach does not fully resolve task interference issues, as the model still requires co-training on both visual-text and robot data. We introduce the mixture-of-expert to resolve the problem, which is in Figure~\ref{fig:framework}. Specifically, given $x^l$ be the input of the $l$-th block. The input can either belong to the $D_{robot}$ or $D_{v-l}$. Notably, we design a dual router, the one to deal with tasks regarding multimodal understanding and conversational ($f(\text{FFN}_{v-l})$), and the other learn representation on robot control ($f(\text{FFN}_{robot})$). The input is first coming through a multi-head self-attention $x^{l'} = \text{MHA}(x^{l-1})+x^{l-1}$, where $\text{MHA}(\cdot)$ represents multi-head self attention. It is then fed into the mixture-of-expert layer, which can be represented as:
\begin{align*}
MoE(x^{l'})&=\begin{cases}
f(\text{FFN}_{v-l})(x^{l'}),&m = 0\\
f(\text{FFN}_{robot})(x^{l'}),&1\leq m\leq M_r
\end{cases}
\end{align*}
This is then added with input from skip connection $x^{l} = x^{l'} + MoE(x^{l'})$. Notice that in stage 1 training, only the control expert is activated. 

To differentiate task outputs, we employ distinct system prompts, such as ``Answer based on question" for understanding and conversation tasks, and ``Predict robot action" for control tasks. Intuitively, a static MoE architecture applied to the MLP layers can be viewed as a high-dimensional feature extractor that partitions the shared parameter space. This allows each task (e.g., understanding and control) to utilize a substantial portion of dedicated neurons, enabling the model to excel at both. A key advantage of this MoE-like architecture is that during inference, only one path is activated, preserving the model parameters of the base model. Our results demonstrate that this straightforward approach leads to simultaneous improvements in understanding, conversation, and control performance.

\begin{figure*}[t]
   \centering
   \includegraphics[width=0.85\linewidth]{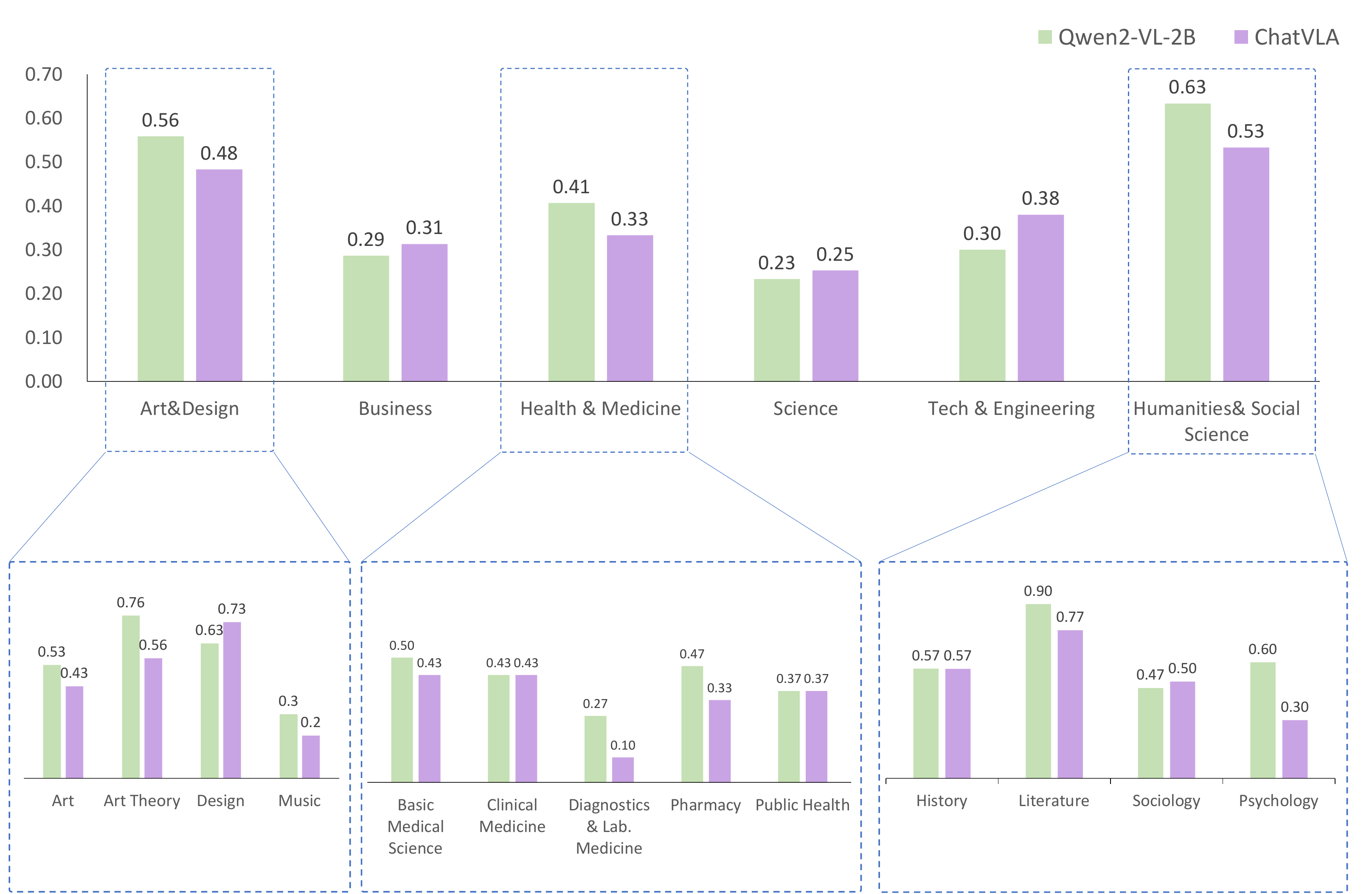}
   \caption{Comparison with Qwen2-VL on $MMMU_{val}$.}
   \label{fig:mmmu_result}
\end{figure*}

\textit{Why sharing self-attention layers?} A prevailing solution is a use mixture of attention to learn task-specific representation. However, based on our experiments (detailed in Section~\ref{sec:exp}), we believe that understanding and robot control tasks share representations that are beneficial to both. For example, a typical robot control scenario requires the model to understand the scene, recognize objects, determine their locations, and then translate this information into actions.  These high-dimensional representations share similar semantic concepts. Therefore, the interconnected nature of these two tasks is crucial for simultaneously improving performance on both understanding and control.

\section{Experiment}
\label{sec:exp}
In this section, we conduct a series of experiments to evaluate the performance of ChatVLA across a range of embodied control and multimodal understanding tasks.

\subsection{Results on Multimodal Understanding and Visual-Question Answering}

We evaluate the visual question answering abilities of ChatVLA using Vlmevalkit~\cite{duan2024vlmevalkit}  on TextVQA ~\cite{singh2019towards}, DocVQA~\cite{mathew2021docvqa}, InfoVQA~\cite{9706887}, AI2D~\cite{kembhavi2016diagram}, ChartQA~\cite{masry-etal-2022-chartqa}, MTVQA~\cite{tang2024mtvqa}, and RealworldQA~\cite{RealWorldQA}.
We also tested against more challenging benchmarks designed for MLLMs, i.e.,
MMMU~\cite{yue2023mmmu}, MMStar~\cite{chen2024we}, MME~\cite{mme}, OCRBench~\cite{Liu_2024}, HallBench~\cite{Guan_2024_CVPR} and MMBench~\cite{mmbench}. 
As delineated in Table \ref{tbl:qa_table}, ChatVLA demonstrates competitive performance relative to existing VLMs across multiple benchmarks. Notably, in VQA tasks, our framework achieves a notable performance of 71.2 on TextVQA, surpassing current SOTA VLAs by substantial margins. Specifically, it outperforms ECoT and DiVLA by relative improvements of 9.2x and 9.5x over these baseline models. The model exhibits particularly strong capabilities in multimodal reasoning tasks requiring complex cross-modal integration. On the MMStar benchmark, ChatVLA attains a score of 37.4, demonstrating 2.2x and 6.9x performance enhancements over DiVLA and ECoT respectively.

\subsection{Results on Real Robot Tasks}

The embodied control performance of ChatVLA is evaluated on 25 realworld manipulation tasks. All these evaluated tasks can be divided into three categories according to the granularity of the language instructions. A more detailed description of these tasks can be found in the Appendix (Section \ref{sec:appendix}). We conducted 528 trials on a real robot to evaluate the model's ability.

\textbf{Long-horizon tasks with direct prompting.} The model is asked to executing tasks directly from language instruction(e.g., ``Sort toys"). The four tasks we evaluated were all completed within a toy scenario constructed on a desktop setup.
Challenging tasks of this category include Task 1, where all toys are randomly positioned in varying poses, and Task 3, which demands the integration of three distinct skills: opening, picking, and closing.
Our method demonstrates substantial advantages in executing tasks directly from high-level descriptions across all evaluated scenarios. The approach maintains consistent performance in multi-step sequences, achieving a 0.54 average success length in Task 1 (6.75x higher than Octo) and perfect success rates throughout Task 3’s three-step sequence.

\textbf{Long-horizon tasks with high-level planner.} The model receives intermediate commands that specify the current sub-task objectives (e.g., ``pick object and place to target location"). The primary challenge in this evaluation stems from the substantial variations between sub-tasks, which involve: (1) diverse object types (e.g., plates, cups, bread), (2) multiple required skills (e.g., pick-place,flip), (3) varying location heights (e.g. top/bottom shelf positions) as visually demonstrated in the bottom-right panel of Fig.\ref{fig:firstfig}. These variations collectively create a testbed for evaluating the model's compositional reasoning capability - specifically, its capacity to integrate object manipulation, spatial reasoning, and interference adaptation. This requirement is clearly reflected in the experimental results shown in Table \ref{tbl:taskscore_compositional}, where our method outperforms OpenVLA and Octo across all task configurations.

\textbf{Cross-skill multi-tasking.} These tasks require the integration of multiple manipulation skills (e.g., picking, placing, pushing, and hanging) across various real-world environments, specifically categorized into three test domains: bathroom scenarios (Tasks 14-17), kitchen environments (Tasks 18-19), and tabletop configurations (Tasks 20-25). As demonstrated in Table \ref{tbl:taskscore_other}, ChatVLA achieves superior performance compared to both Octo and OpenVLA across all task categories. The model exhibits particularly strong performance in challenging bathroom and kitchen tasks, where robotic arm operations are constrained to a severely limited spatial range. This experimental setup inherently introduces substantial safety considerations during model evaluation, consequently establishing rigorous requirements for the operational precision and system robustness of the assessed models.

\subsection{Ablation Study}

\textbf{What vision-language data are preferred?}
In stage 2, we employed the LLaVA-1.5~\cite{llava1.5} dataset for co-training, which allowed the model to achieve compatible results on both VQA and MLLM benchmarks compared to Qwen2-VL. However, we argue that the remaining performance gap is attributed to the limitations of the visual-text data used. To explore this further, we conducted an in-depth analysis of the results between ChatVLA and Qwen2-VL on the MMMU dataset, as illustrated in Fig. \ref{fig:mmmu_result}.

The MMMU dataset is divided into six categories, and ChatVLA's performance is slightly lower than Qwen2-VL in three of them: art, medicine, and social science. A closer inspection of the results for the corresponding subcategories reveals that the performance discrepancies primarily occur in five specific domains: art theory, lab medicine, pharmacy, literature, and psychology. These fields involve relatively limited specialized knowledge that is difficult to obtain. Upon reviewing the composition of the LLaVA dataset, we were surprised to find that its subdatasets, including COCO, GQA, OCR-VQA, TextVQA, and VisualGenome, lack the expert knowledge required for these domains, which likely contributed to the observed performance drop. A more detailed description of the composition of these datasets can be found in the Section~\ref{sec:llava_dataset}.

This finding also highlights the considerable potential of our model: with more appropriate professional data for training, we believe that we can achieve significantly better performance in multimodal understanding.

\textbf{What is the appropriate ratio of visual-text data to robot data?}
While co-training with visual-text data, we followed the settings discussed in ECoT~\cite{ecot} and set the overall visual-text data to robot data ratio at 1:3. However, whether other data ratios are beneficial or detrimental to multimodal understanding and robot tasks still requires attention. Therefore, under the same number of steps, we modified the ratio of visual-text data to robot data in co-training to 1:1 and 3:1, respectively. The results of the three setups are shown in the Table~\ref{tbl:ratio}.
\begin{table*}[tb]
  \centering
  \caption{\textbf{Understanding task.} Evaluation of MLLMs and VLAs on 6 multimodal understanding benchmarks and 7 VQA benchmarks. We use bold to denote top-ranked methods, and underlined entries signify secondary performers.} 
  \label{tbl:ratio}
  \resizebox{1.0\linewidth}{!}{
      \begin{tabular}{c|ccccc|ccccccc}
        \toprule
        \multirow{2}{*}{Method} & \multicolumn{5}{c|}{Multimodal Understanding Benchmarks} & \multicolumn{7}{c}{VQA Benchmarks} \\
               & MMMU & MMStar & MME & OCRBench & HallBench  
                & TextVQA & DocVQA & InfoVQA & AI2D & ChartQA & MTVQA & RealWorldQA \\
        \midrule
        
        1:1& 36.1&44.7&1426.9&691&36.2&72.6&82.9&54.0&65.382&62.6&10.0&57.9\\
        3:1&35.3&45.3&1399.5&726&36.4&72.7&83.6&54.3&67.0&63.2&10.3&58.8\\
       
      1:3 & 37.4 &47.2 & 1435.2 & 729 & 39.9 & 71.2 & 83.3 & 53.3 & 67.6 & 59.9 & 11.5 & 57.0 \\
        \bottomrule
      \end{tabular}
  }
\end{table*}

Surprisingly, a smaller amount of visual-text data resulted in better performance. This aligns with the analysis in the previous subsection~\ref{sec:motivation} in the paper, which suggests that even a limited amount of visual-text data is sufficient to reactivate visual-text alignment and bridge the real-world interacting capacity gap between the base VLM and the VLA model.

\section{Conclusion}
Integrating embodied control and multimodal understanding in Vision-Language-Action (VLA) models is challenging, as current methods often compromise one for the other. We identified key limitations: robot-only training degrades conversational ability, while visual-text co-training diminishes control performance due to spurious forgetting and task interference. To address this, we introduce ChatVLA, a unified framework combining Phased Alignment Training (prioritizing control before multimodal linking) and a Mixture-of-Experts architecture. ChatVLA achieves competitive VQA and general understanding performance while excelling at real-world robot control (25 tasks across diverse scenes), outperforming OpenVLA and ECoT with 3.5x fewer parameters. These results demonstrate that a single network can effectively harmonize multimodal reasoning, conversation, and physical interaction.

{
    \small
    \bibliographystyle{ieeenat_fullname}

}

\clearpage
\setcounter{page}{1}
\maketitlesupplementary

\section{Appendix}
\label{sec:appendix}

\subsection{Implementation Details}

\textbf{Data details.}
For visual-text data, we use llava-1.5~\cite{llava1.5} dataset for co-training. Following the data ratio mentioned in ECOT, we use set the ratio of visual-text data to robot data as 1:3. Using robot data, we evaluated our method on 25 real-world robot tasks, including long-horizon tasks with direct prompting. The data was randomly sampled from the LLaVA fine-tuning dataset.  We hypothesize that carefully curated data is crucial for mitigating spurious forgetting, a topic we plan to explore in future work.

\textbf{Training details.} We use Qwen2-VL-2B as our VLM backbone and the set of action head follows DiVLA~\cite{wen2024diffusionvla}.
We train our ChatVLA using a phased alignment training, as is discussed in Section \ref{sec:architecture_data}. In the first stage, we train our model on robot data, only activating the control expert and its corresponding action head. In the second stage, we co-train both visual-text data and robot data. Both control expert and understanding expert are trained using the same learning rate of 2e-5.

\textbf{Evaluation metrics.}
The calculation for long-horizon tasks in Table ~\ref{tbl:taskscore_longhorizon} and Table ~\ref{tbl:taskscore_compositional} is as follows: One point is awarded for each successfully completed step. Once all steps are executed, the total score is computed. ``Avg. Len." represents the average success length, which is calculated by recording the lengths of successful sequences across multiple executions of long-horizon tasks and then computing its average. This provides a key indicator of the model's performance in handling long-horizon tasks based on the length of successful operation sequences.

For real robot cross-skill multi-tasking in Table~\ref{tbl:taskscore_other},``Avg." denotes the average success rate across all these tasks, calculated by dividing the total number of successful task executions by the total number of test executions for these tasks.

\textbf{Dataset description.}\label{sec:llava_dataset}
The visual-text pairs in LLaVA are composed of:
\begin{itemize}
    \item COCO: Contains images annotated with object labels, segmentations, and captions, supporting tasks like detection, segmentation, and image captioning.
    \item GQA: Focuses on visual reasoning, with images paired with questions that require logical interpretation of visual content.
    \item OCR-VQA: Combines visual question answering with OCR, emphasizing images that contain textual information needing extraction and reasoning.
    \item TextVQA: A visual question answering dataset where answers are derived from reading and understanding text embedded in images.
    \item VisualGenome: Provides detailed annotations of objects, attributes, relationships, and regions in images, supporting scene understanding and visual question answering.
\end{itemize}

\subsection{Robot task}

\begin{figure}[t]
   \centering
   \includegraphics[width=\linewidth]{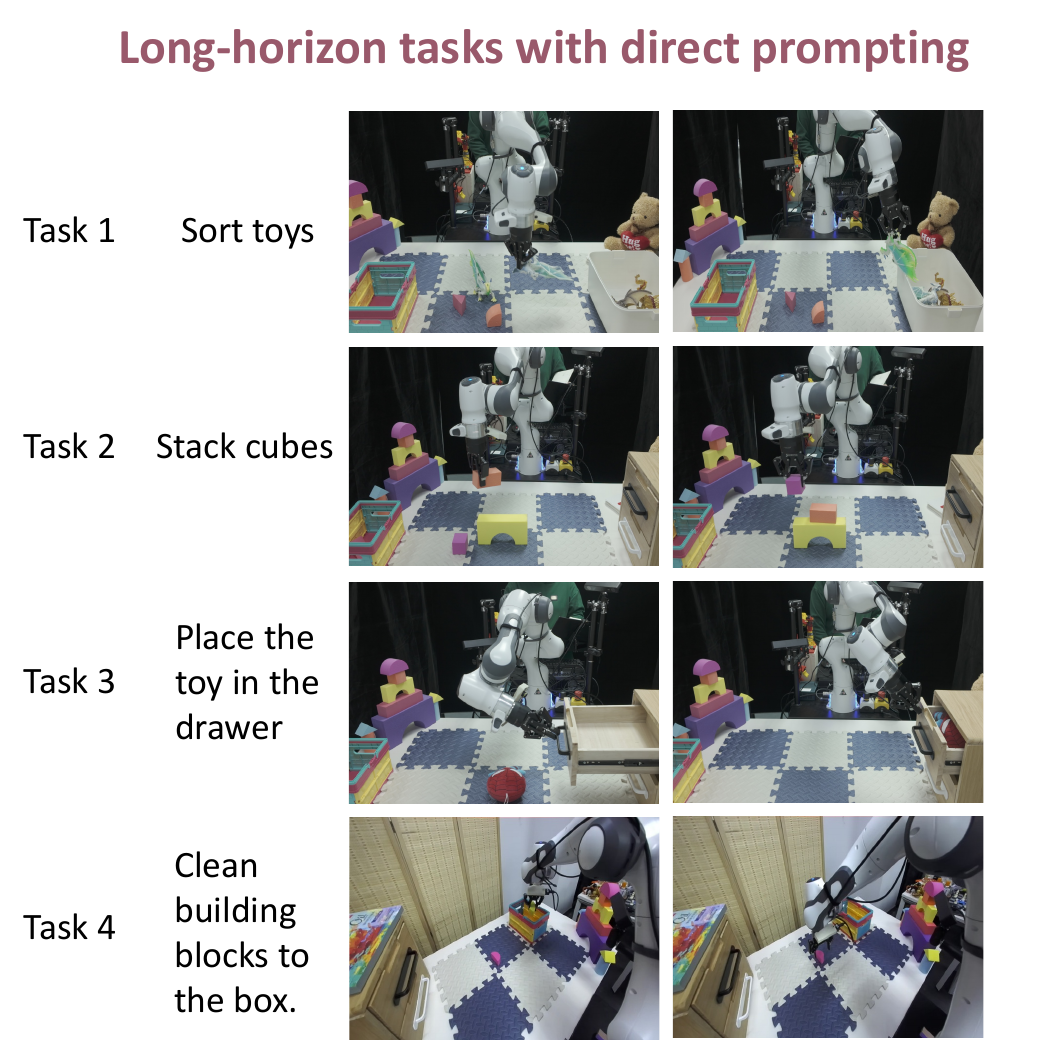}
   \caption{Settings of Long-horizon tasks with direct prompting}
   \label{fig:realpic_direct}
\end{figure}

\begin{figure*}[t]
   \centering
   \includegraphics[width=\linewidth]{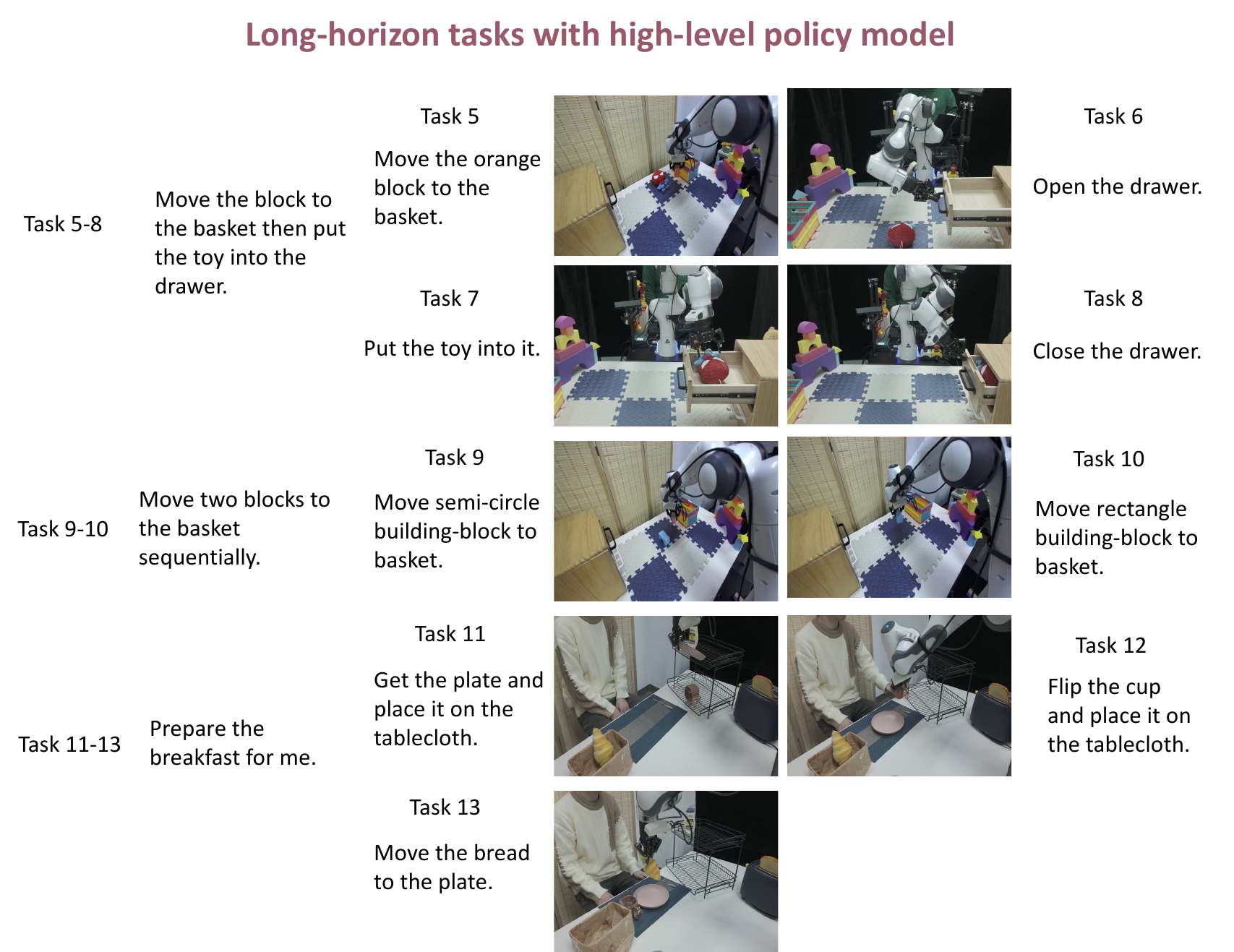}
   \caption{Settings of Long-horizon tasks with high-level planner}
   \label{fig:realpic_highlevel}
\end{figure*}

\begin{figure*}[t]
   \centering
   \includegraphics[width=\linewidth]{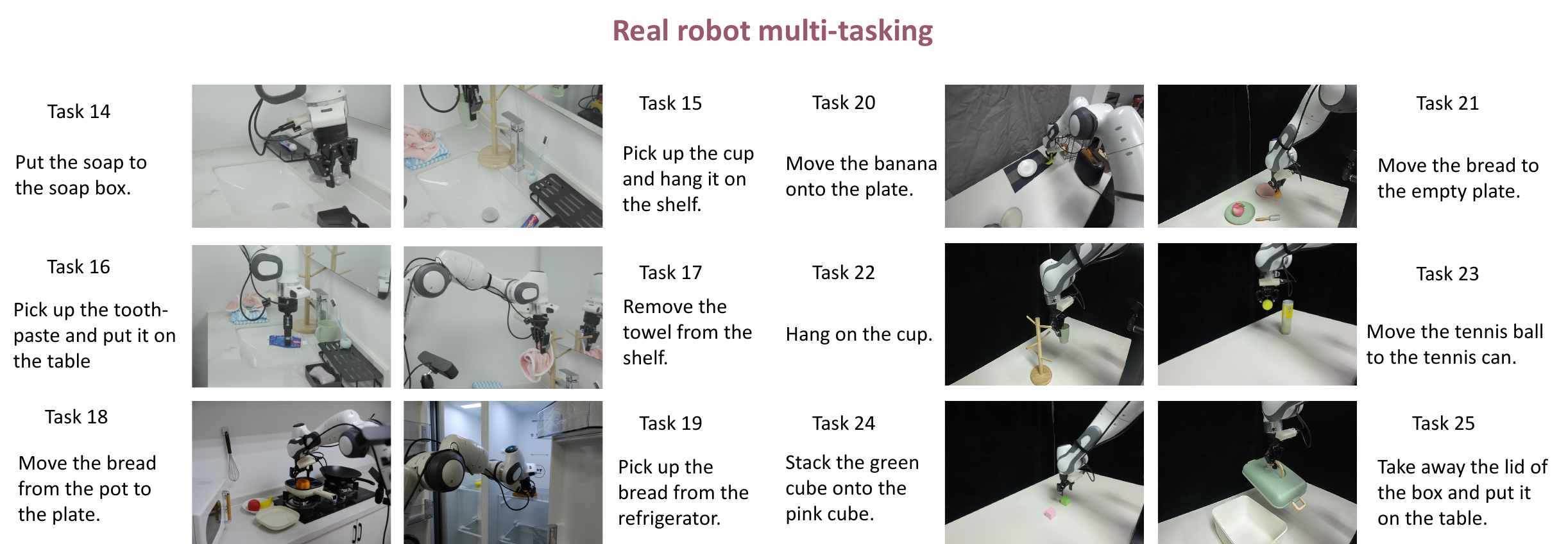}
   \caption{Settings of Cross-skill real robot multi-tasking.}
   \label{fig:realpic_other}
\end{figure*}

The embodied control performance of ChatVLA is evaluated on 25 real world manipulation tasks.

\textbf{Long-horizon tasks with direct prompting.} As is shown in  \ref{fig:realpic_direct}, all the tasks of this category are set under a real world toy scene.
\begin{itemize}
    \item Task 1: Sort toys. On the desktop, there are two toy animals with random positions and postures, as well as two building blocks. The robotic arm needs to place all the animals on the desktop in the box on the left and all the building blocks in the basket on the right.
    \item  Task 2: Stack cubes. The robotic arm first needs to pick up the orange building block from the right side and stack it on the yellow building block in the middle. Then, it needs to pick up the smallest pink square and stack it on the orange building block that was just stacked.  
    \item  Task 3: Place the toy in the drawer. The drawer is closed. Therefore, the robotic arm first needs to rotate and pull open the drawer. Then, it should pick up the toy on the table and place it into the drawer. Finally, close the gripper to shut the drawer.
    \item  Task 4: Clean building blocks to the box.
 The robotic arm needs to put the building blocks on the table into the box on the right side one by one until there are no more building blocks on the table. 
\end{itemize}

\textbf{Long-horizon tasks with high-level planner.} The settings are shown in  \ref{fig:realpic_highlevel}. 

\begin{itemize}
    \item Task 5: Move the orange block to the basket. The robotic arm needs to pick up the building block next to the doll on the table and place it into the box on the right side. 
    \item Task 6: Open the drawer. The robotic arm needs to rotate and grip the drawer handle, and then move parallel to the right to open the drawer. 
    \item Task 7: Put the toy into it. The robotic arm needs to pick up the toy in the middle and place it into the open drawer. 
    \item Task 8: Close the drawer. The robotic arm needs to close the gripper and gently push the open drawer to the left until the drawer is closed. 
\end{itemize}

\begin{itemize}
    \item Task 9: Move semi-circle building-block to basket. The robotic arm needs to pick up the semi-circular building block and place it into the basket on the right side. 
    \item Task 10:Move rectangle building-block to basket. The robotic arm needs to pick up the rectangle building block and place it into the basket on the right side. 
\end{itemize}

\begin{itemize}
    \item Task 11: Get the plate and place it on the tablecloth. The robotic arm needs to pick up the pink plate from the upper part of the shelf on the right side and then place it on the tablecloth at the center of the table.  
    \item Task 12: Flip the cup and place it on the tablecloth. The robotic arm needs to go to the bottom layer of the shelf on the right side, grip the mug, then turn it over and place it on the tablecloth in the middle of the table. 
    \item Task 13: Move the bread to the plate. The robotic arm needs to grip the bread from the bread basket on the left side and place it on the plate that was just taken down. 

\end{itemize}

\textbf{Cross-skill multi-tasking.} The settings are shown in  \ref{fig:realpic_other}.
\begin{itemize}
    \item  Task 14:Put the soap to the soap box. This is a bathroom task. The robotic arm needs to pick up the soap from the left side of the washbasin and place it into the soap dish on the right side of the washbasin. 
    \item  Task 15:Pick up the cup and hang it on the shelf. This is a bathroom task. The robotic arm needs to pick up the cup from the sink and hang it on the shelf in front of the mirror. 
    \item  Task 16:Pick up the tooth-paste and put it on the table. This is a bathroom task. The robotic arm needs to pick up the toothpaste from the sink and place it on the table.
    \item  Task 17:Remove the towel from the shelf. This is a bathroom task. The robotic arm needs to take down the towel hanging on the shelf and place it on another towel. 
    \item Task 18:Move the bread from the pot to the plate. This is a kitchen task. The robotic arm needs to pick up the bread from the pot and place it on the plate. 
    \item Task 19:Pick up the bread from the refrigerator. This is a kitchen task. The robotic arm needs to find the bread in the refrigerator and pick it up. 
    \item Task 20:Move the banana onto the plate. The robotic arm needs to pick up the banana at a random position and place it on the plate in the middle. 
    \item  Task 21: Move the bread to the empty plate. The robotic arm needs to ignore the distractions, grip the bread, and then find the empty one among the two plates in front of it, and put the bread into that plate. 
    \item  Task 22:Hang on the cup. The robotic arm needs to pick up the mug and hang it on the shelf on the left side. 
    \item Task 23:Move the tennis ball to the tennis can. The robotic arm needs to pick up the tennis ball and lift it up to place it into the tennis ball can. 
    \item Task 24:Stack the green cube onto the pink cube. The robotic arm needs to pick up the green cube on the right and stack it on top of the square on the left side. 
    \item  Task 25:Take away the lid of the box and put it on the table. The robotic arm needs to pick up the lid that is covering the box on the left side of the table and place the lid on the tabletop in the middle. 

\end{itemize}

\end{document}